\documentclass{article}
\usepackage{spconf,amsmath,graphicx,hyperref}

\usepackage{cite}
\usepackage{amsmath,amssymb,amsfonts}
\usepackage{algorithmic}
\usepackage{graphicx}
\usepackage{textcomp}
\usepackage{xcolor}
\usepackage{amssymb}
\usepackage[capitalize]{cleveref}
\usepackage{color}
\usepackage{booktabs} 
\usepackage{url}
\usepackage{algorithm}
\usepackage{algorithmic}
\usepackage{tabularx}   %
\usepackage{booktabs}
\usepackage{tikz}
\usepackage{pgfplots}
\pgfplotsset{compat=1.18}

\title{Virtual Fashion Photo-Shoots:\\Building a Large-Scale Garment-Lookbook Dataset}
\name{Yannick Hauri \qquad Luca A. Lanzendörfer \qquad Till Aczel}
\address{ETH Zurich}
\begin{document}
\ninept
\maketitle
\begin{abstract}
Fashion image generation has so far focused on narrow tasks such as virtual try-on, where garments appear in clean studio environments.  
In contrast, editorial fashion presents garments through dynamic poses, diverse locations, and carefully crafted visual narratives.  
We introduce the task of virtual fashion photo-shoot, which seeks to capture this richness by transforming standardized garment images into contextually grounded editorial imagery.  
To enable this new direction, we construct the first large-scale dataset of garment–lookbook pairs, bridging the gap between e-commerce and fashion media.  
Because such pairs are not readily available, we design an automated retrieval pipeline that aligns garments across domains, combining visual–language reasoning with object-level localization.  
We construct a dataset\footnote{https://huggingface.co/datasets/disco-eth/lookbook} with three garment–lookbook pair accuracy levels: high quality (10,000 pairs), medium quality (50,000 pairs), and low quality (300,000 pairs).  
This dataset offers a foundation for models that move beyond catalog-style generation and toward fashion imagery that reflects creativity, atmosphere, and storytelling.
\end{abstract}

\begin{keywords}
Virtual Photo-Shoot, Dataset Curation, Fashion Image Generation, Garment-Lookbook Pairs, Image Retrieval
\end{keywords}

\section{Introduction and Related Work}\label{sec:introduction}

Advances in image generation and the growth of the fashion industry have driven research in virtual try-on, fashion image editing, clothing recognition, and garment classification.  
Virtual try-on, in particular, allows users to upload an image and generate how different garments would appear when worn.   

Building on virtual try-on, we introduce the task of \textit{virtual photo-shoot}, which aims to generate editorial-style images of models wearing a given garment in diverse, complementary settings.  
This enables designers and fashion houses to automatically produce creative photo-shoot material, moving beyond the studio-like outputs of existing try-on systems.  

\begin{figure}[t]
    \centering
    \includegraphics[width=\columnwidth]{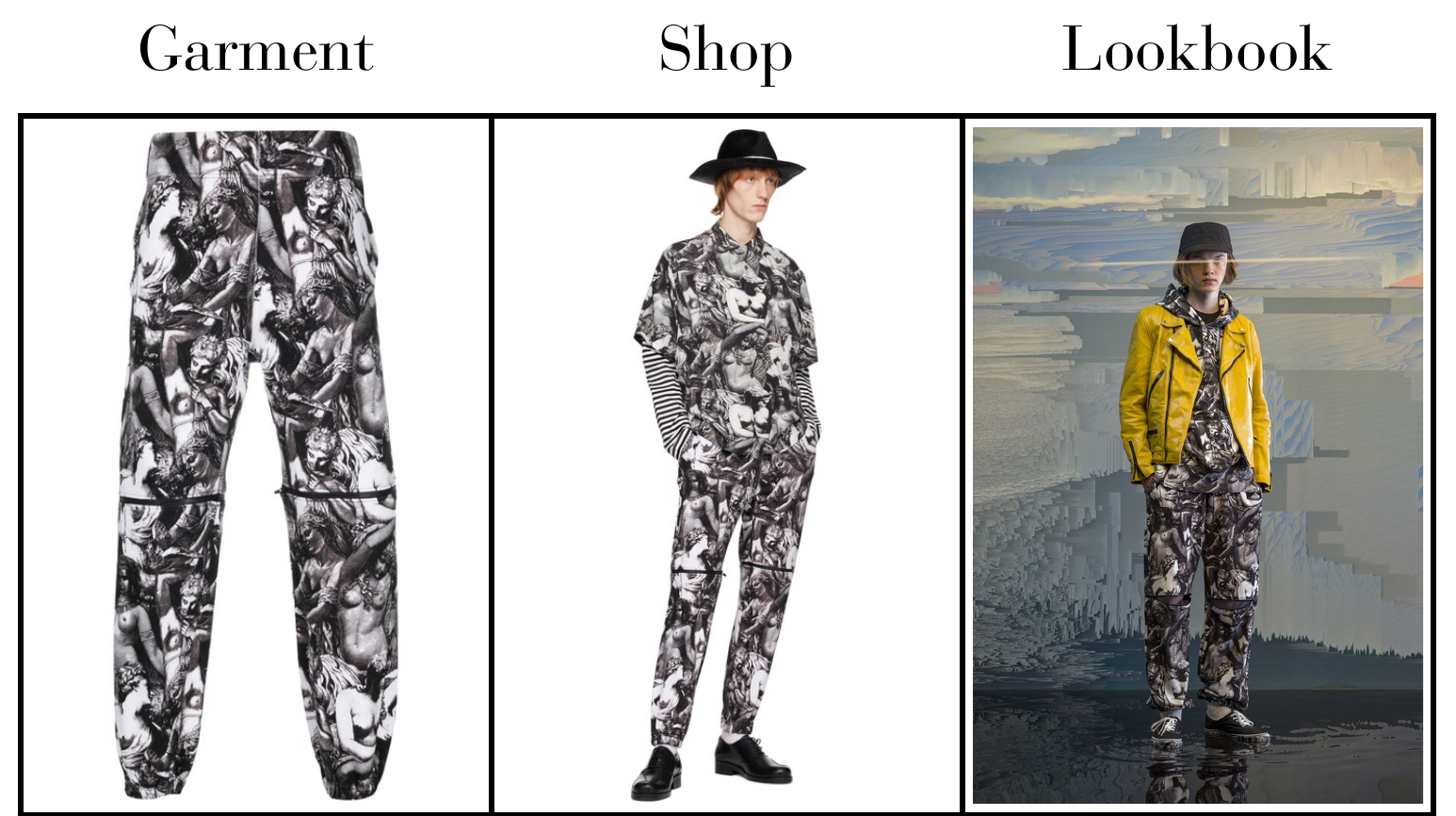}
    \caption{Difference in garment, shop, and lookbook image. Existing datasets provide clean shop images, not suitable for virtual photoshoot model training.}
    \label{shop-lookbook-fig}
\end{figure}

Training such models requires data linking garment-level product images with lookbook-style photography. 
Existing datasets~\cite{liu_deepfashion_2016,han_viton_2018,ge_deepfashion2_2019,choi_viton-hd_2021,morelli_dress_2022} provide rich annotations but focus on shop environments. 
Figure~\ref{shop-lookbook-fig} illustrates the difference between shop and lookbook style images.  
These datasets pair isolated garment images (with uniform white backgrounds) with shop-lookbook images, which exhibit minimal variation in poses, backgrounds, and styling.  
Consequently, current datasets do not capture the creative diversity of real fashion media.

To address this gap, we construct the first dataset of garment–lookbook pairs.  
In this setting, the garment image provides a standardized product-level reference, while the lookbook image captures the same garment in diverse poses, backgrounds, and artistic styles.  
By linking these two domains, the dataset enables training models that can generate lookbook-style photographs conditioned on a garment image, analogous to how virtual try-on models generate studio-style outputs from garment–shop pairs.  
Unlike try-on datasets, which can be collected directly from e-commerce product pages, garment–lookbook pairs are not co-located and must be assembled from separate sources.  
We therefore gather unpaired garment and lookbook images from diverse collections and create pairs automatically through garment retrieval.  

\begin{figure*}[t]
    \centering
    \includegraphics[width=0.32\textwidth]{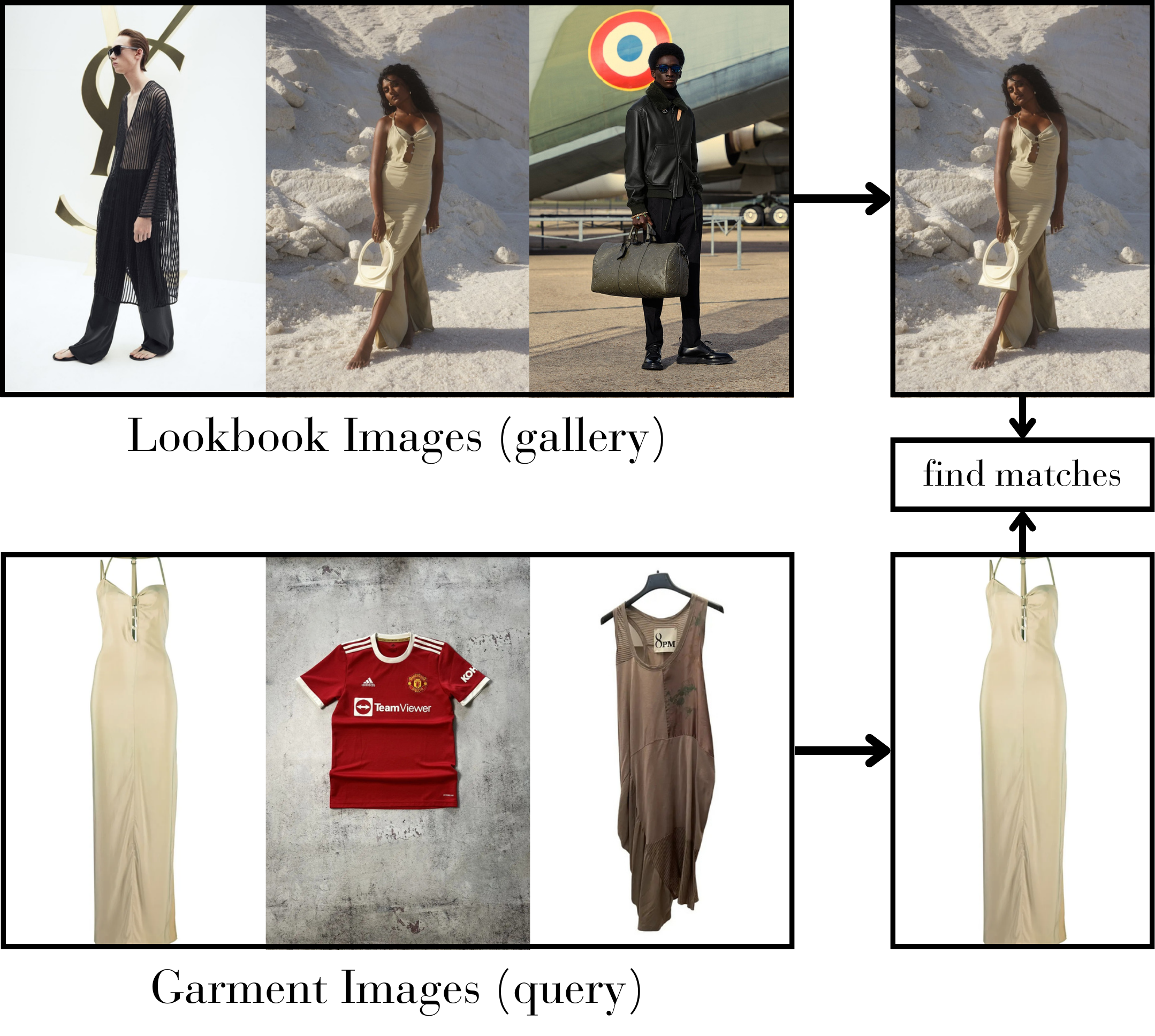}%
    \hfill
    \raisebox{0.15\height}{\includegraphics[width=0.65\textwidth]{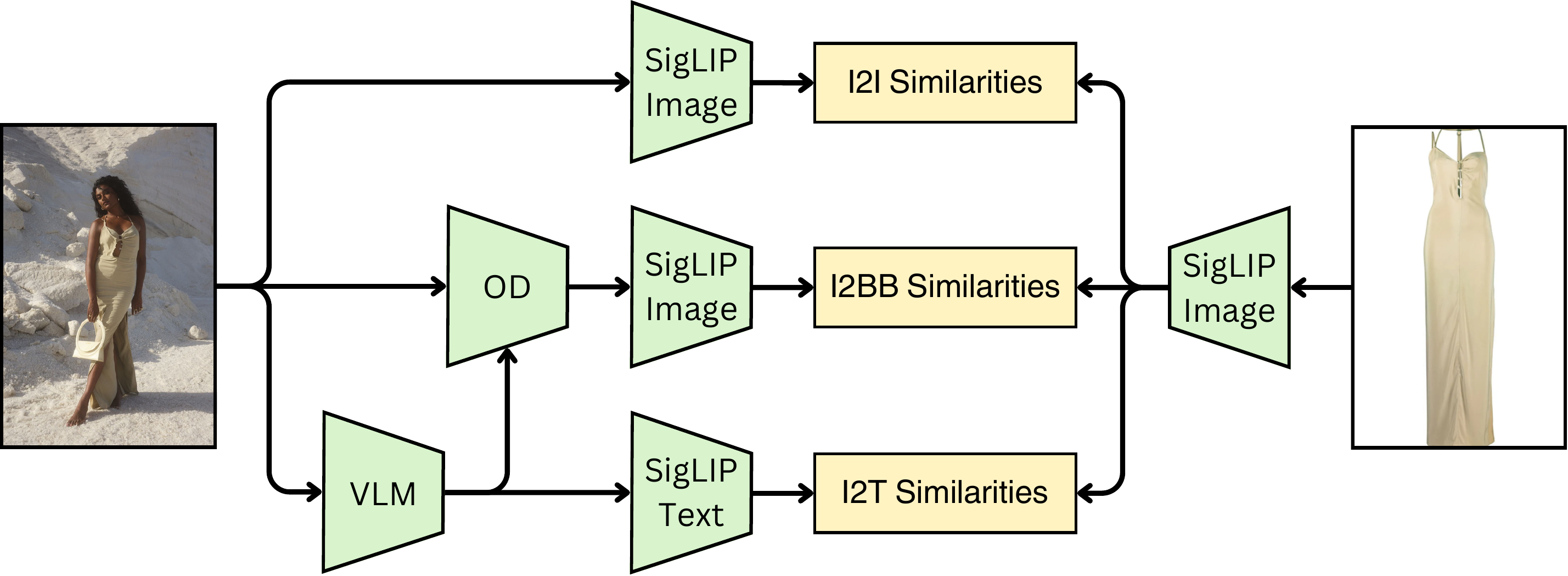}}
    \caption{\textbf{Left}: Examples of lookbook images (gallery) and garment images (query), showing a sample where one garment appears in both a gallery and a query image. Matching gallery-query pairs form the basis of our dataset, and the task is to find all such matches. \textbf{Right}: Overview of our retrieval pipeline. Query images are embedded with SigLIP2, while garment descriptions for gallery images are generated with a vision-language model (VLM). Object detection (OD) pconditioned on the garment description roduces bounding boxes for individual garments. Embeddings of gallery images, descriptions, and bounding boxes are compared with SigLIP2 to compute image-to-image, image-to-bbox, and image-to-text similarities.}
    \label{merged-fig}
\end{figure*}

Several existing retrieval approaches have been applied to fashion matching.
Proxynca++~\cite{vedaldi_proxynca_2020} leverages proxy-based contrastive learning to capture fine-grained visual similarity, enabling accurate image-to-image matching in structured datasets.
Hyp-DINO~\cite{ermolov_hyperbolic_2022} encodes hierarchical embeddings in hyperbolic space, capturing relationships between visually similar garments more effectively than standard Euclidean embeddings.
While these models perform well on clean, curated datasets, they are less robust to the diverse backgrounds, poses, and editing styles found in editorial fashion imagery.

To overcome these limitations, we develop a retrieval pipeline combining vision–language models (VLMs), object detection (OD), and SigLIP-based similarity estimation~\cite{tschannen_siglip_2025}.
VLMs identify garment categories in natural language, OD isolates relevant regions in lookbook images, and SigLIP provides robust similarity scores between garment crops and query images.
This combination is particularly effective in noisy, heterogeneous settings, where existing metric-learning models like Proxynca++ and Hyp-DINO alone struggle.

By integrating these complementary approaches into an ensemble, we improve retrieval accuracy and robustness across diverse datasets containing complex poses, backgrounds, and editorial styles.
This high-quality garment–lookbook matching enables the construction of a dataset suitable for training generative models to produce rich, contextually grounded virtual photo-shoots.

In summary, this work makes three contributions:  
(1) We define the new task of virtual photo-shoot and present the first large-scale dataset of garment–lookbook pairs.
(2) We propose a zero-shot retrieval pipeline integrating VLMs, OD, and SigLIP for automatic garment–lookbook matching.
(3) We provide an ensemble-based retrieval strategy that further improves the quality of paired data.

\section{Methodology}

To support the task of virtual photo-shoot, we construct a large-scale dataset of garment–lookbook pairs linking product-level garment images with lookbook images.  
The dataset is built in two stages: first, we collect unpaired garment and lookbook images; then, we create pairs via retrieval.  
This approach reflects the reality that product pages and editorial media are rarely co-located for most brands.  
A visualization of the retrieval stage is shown in Figure~\ref{merged-fig}.

\subsection{SigLIP2 Retrieval}
\label{sec:siglip-retrieval}

Garment–lookbook retrieval is challenging because lookbook images contain multiple garments, diverse poses, and complex backgrounds, while product-level garment images are clean and standardized.  
Directly embedding the garment and full lookbook image produces suboptimal matches.
We refer to this baseline as \textbf{SigLIP2-FI2I} (Full Image-to-Image).

To improve retrieval, we introduce \textbf{SigLIP2-T2I} (Text-to-Image).  
A vision–language model (gpt-4.1-mini~\cite{bahrini_chatgpt_2023}) parses each lookbook image into individual garment components and generates concise natural language descriptions.  
We then compute similarity between the product garment’s SigLIP2 image embedding and the SigLIP2 text embedding of each description, improving robustness by filtering out background clutter and focusing on garment content.

Relying only on text, however, discards fine visual cues such as patterns, texture, and subtle design elements.  
To recover these, we propose \textbf{SigLIP2-BB2I} (Bounding Box-to-Image), where an open-vocabulary object detector (YOLO-World~\cite{cherti_reproducible_2023}), guided by the text descriptions, predicts bounding boxes for each garment in the lookbook.  
We crop these regions, embed them with SigLIP2, and compare them to the product garment embedding, yielding localized image-to-image similarities.

Since SigLIP2-BB2I produces multiple scores per lookbook image, we aggregate them with the full-image similarity from SigLIP2-FI2I by taking the maximum.  
This final strategy, \textbf{SigLIP2-I2I} (Image-to-Image), leverages both global and localized cues while avoiding dilution by weaker matches.

\subsection{Ensemble Retrieval}
\label{sec:ensemble-retrieval}

While the SigLIP2 pipeline provides strong zero-shot performance, further gains can be achieved by incorporating complementary retrieval models.  
We therefore extend our approach into an ensemble retrieval system that combines SigLIP2 with specialized metric learning methods.
Specifically, we include two state-of-the-art distance metric learning models: Proxynca++ \cite{vedaldi_proxynca_2020} and Hyp-DINO \cite{ermolov_hyperbolic_2022}.  
These models capture garment-specific details and structural cues beyond what SigLIP2 alone provides, producing similarity scores that complement the SigLIP2-based similarities.

Because the similarity distributions of different models are not directly comparable, we normalize them before combining.  
For each model $m$, we estimate the mean $\mu_m$ and standard deviation $\sigma_m$ of its similarity scores.  
We then transform each score $s_{mij}$ for query–gallery pair $(i, j)$ into a standardized score
\begin{equation}
    s'_{mij} = \frac{s_{mij}-\mu_m}{\sigma_m},
\label{normalize-eq}
\end{equation}
so that all models operate on a common standardized scale.

With this normalization, we combine similarities from multiple models.  
Merging SigLIP2-I2I and SigLIP2-T2I yields the \textbf{SigLIP2-Ensemble}, while combining all four models (SigLIP2-I2I, SigLIP2-T2I, Proxynca++, Hyp-DINO) produces the \textbf{Total-Ensemble}.

\subsection{Dataset}
\label{sec:dataset-methodology}

We collect approximately 550,000 lookbook and runway images with associated metadata from SHOWstudio\footnote{https://www.showstudio.com/} and Tagwalk\footnote{https://www.tag-walk.com/}.  
We augment these collections with roughly 9.5 million garment images from e-commerce platforms such as Farfetch\footnote{https://www.farfetch.com}, VestiaireCollective\footnote{https://us.vestiairecollective.com/}, Grailed\footnote{https://www.grailed.com/}, and Depop\footnote{https://www.depop.com/}, using brand names from the editorial metadata as queries.  
Metadata such as brand name and short descriptions are retained to assist search and filtering.  
The resulting corpus contains about ten million images and, to our knowledge, represents the first large-scale resource tailored to the virtual photo-shoot task.  
We include runway images under the lookbook category as a practical compromise, since runway photography contributes editorial diversity that is rarely available at scale from single-brand lookbooks.  

Pairing images follows the methodology described in Sections~\ref{sec:ensemble-retrieval}.  
For each query garment, we compare its brand name with those of gallery candidates using fuzzy string matching (RapidFuzz~\cite{rapidfuzz}).  
Only gallery images with sufficiently similar brand names are retained.  
Among these candidates, we select the lookbook image with the highest ensemble similarity score to form a garment–lookbook pair.  
Sorting all pairs by similarity produces a curated dataset aligned with both visual and semantic consistency.

To create quality splits, we rank each garment image by its highest similarity score.  
The top 10,000 pairs form the \emph{high-quality} set, the top 50,000 pairs form the \emph{medium-quality} set, and the top 300,000 pairs form the \emph{low-quality} set.

\section{Experimental Setup}

To evaluate retrieval strategies for pairing images in our dataset, we require a benchmark dataset that provides ground truth garment–lookbook pairs.  
We consider four established datasets: DeepFashion In-Shop \cite{liu_deepfashion_2016}, DeepFashion Consumer-to-Shop \cite{liu_deepfashion_2016}, DeepFashion2 \cite{ge_deepfashion2_2019}, and DressCode \cite{morelli_dress_2022}.  
Qualitative inspection of our collected data indicates that DressCode is the closest in style and content, with DeepFashion2 as the next most similar.

Nevertheless, our dataset is over two orders of magnitude larger and far more variable and noisy.  
Lookbook images span diverse backgrounds, poses, and editing effects, while garment presentation varies in quality.  
Although DressCode is simpler than our dataset, it remains the best available proxy for evaluating retrieval performance.  
We report results on DressCode, noting these scores likely overestimate performance on our noisier, more diverse data.  
To train supervised models (Proxynca++ and Hyp-DINO), we use the remaining datasets (DeepFashion In-Shop, DeepFashion Consumer-to-Shop, DeepFashion2).

The SigLIP2-based retrieval models (SigLIP2-I2I, SigLIP2-T2I, and SigLIP2-BB2I) operate in a zero-shot fashion and require no training.
In contrast, the two metric learning based approaches: Proxynca++ \cite{vedaldi_proxynca_2020} and Hyp-DINO \cite{ermolov_hyperbolic_2022}, require training.
Proxynca++ is trained for 80 epochs with 5 warm-up epochs, while Hyp-DINO is trained for 400 epochs.  
The metric standardization uses means and standard deviations estimated on a random subsample of our raw dataset.
For evaluation, we compute query–gallery similarities using the FAISS~\cite{douze_faiss_2025} library.

\begin{table*}[t]
    \centering
    \begin{tabularx}{\textwidth}{l *{9}{>{\centering\arraybackslash}X}}
        \toprule
        & \multicolumn{3}{c}{DeepFashion Shop} 
        & \multicolumn{3}{c}{DeepFashion Consumer} 
        & \multicolumn{3}{c}{DeepFashion2} \\
        \cmidrule(lr){2-4}\cmidrule(lr){5-7}\cmidrule(lr){8-10}
        & R@1 & R@5 & R@10 & R@1 & R@5 & R@10 & R@1 & R@5 & R@10 \\
        \midrule
        Proxynca++       & \textbf{90.0\%} & 96.9\% & 98.0\% 
                         & 33.2\% & 51.1\% & 56.7\% 
                         & 45.2\% & 63.8\% & 71.1\% \\
        Hyp-DINO         & \textbf{90.0\%} & \textbf{97.0\%} & \textbf{98.1\%} 
                         & \textbf{42.5\%} & \textbf{61.1\%} & \textbf{65.9\%} 
                         & 49.9\% & 68.5\% & 74.7\% \\
        SigLIP2-Ensemble (Ours) & 83.6\% & 95.5\% & 97.4\% 
                         & 23.1\% & 40.5\% & 48.0\% 
                         & \textbf{53.6\%} & \textbf{71.8\%} & \textbf{78.8\%} \\
        \bottomrule
    \end{tabularx}
        \caption{Retrieval performance (\%) across three fashion benchmarks. Proxynca++ and Hyp-DINO were trained on gallery-query pairs from the DeepFashion in-shop, DeepFashion consumer-to-shop, DeepFashion2, and DressCode datasets. Note that the SigLIP2-Ensemble was not trained on any of these datasets. The highest recall is marked in bold. Even with training on the datasets, the SigLIP2-Ensemble performs on pair, or outperforms the baselines.}
    \label{tab:retrieval-results}
\end{table*}

\section{Results}

\subsection{Retrival Model}

\begin{figure}[t]
\centering
\begin{tikzpicture}
\begin{axis}[
    scale only axis,
    height=4cm,
    width=4cm,
    colormap={redblue}{rgb255=(0,0,180) rgb255=(255,255,255) rgb255=(180,0,0)},
    colorbar,
    point meta min=-1,
    point meta max=1,
    y dir=reverse,
    enlargelimits=false,
    axis on top,
    xtick={0,...,3},
    ytick={0,...,3},
    xticklabels={SigLIP2-I2I, SigLIP2-T2I, Hyp-DINO, Proxynca++},
    yticklabels={SigLIP2-I2I, SigLIP2-T2I, Hyp-DINO, Proxynca++},
    x tick label style={rotate=45, anchor=east, font=\small},
    y tick label style={font=\small},
]
\addplot [matrix plot*, mesh/cols=4, point meta=explicit,
    nodes near coords,
    every node near coord/.style={
        anchor=center,
        text height=1.5ex,
        text depth=.25ex,
        font=\scriptsize,
        /pgf/number format/fixed,
        /pgf/number format/precision=2
    }
] table [meta index=2] {
x y value
0 0 1.00
1 0 0.48
2 0 0.03
3 0 -0.05
0 1 0.48
1 1 1.00
2 1 0.06
3 1 -0.06
0 2 0.03
1 2 0.06
2 2 1.00
3 2 -0.19
0 3 -0.05
1 3 -0.06
2 3 -0.19
3 3 1.00
};
\end{axis}
\end{tikzpicture}
\caption{Rank correlation heatmap between retrieval models. Values are rounded to two decimals and centered in each cell.}
\label{fig:correlation-heatmap}
\end{figure}
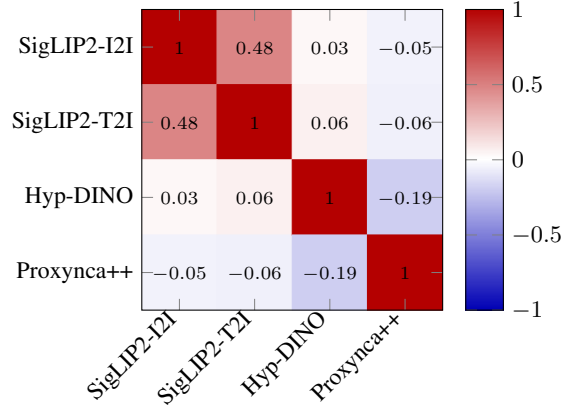

The pairwise correlations between our retrieval models (Figure~\ref{fig:correlation-heatmap}) show that the models are only weakly correlated.
Values range from slightly negative to moderately positive, indicating that each model captures complementary aspects of garment similarity.
This observation motivates our ensemble approach, as combining multiple diverse models allows us to leverage their individual strengths and improve overall retrieval performance.

Table~\ref{tab:retrieval-results} compares Proxynca++, Hyp-DINO, and the SigLIP2-Ensemble.
Proxynca++ and Hyp-DINO are trained on the combined training split, while all models are evaluated on the validation sets.
Despite never seeing these datasets during training, the SigLIP2-Ensemble performs competitively on clean benchmarks and even surpasses metric learning approaches on DeepFashion2.
This advantage likely arises from DeepFashion2 containing more contextual noise around garments, where the VLM+OD pipeline of SigLIP2 effectively isolates relevant features.
To the best of our knowledge, we achieved state-of-the-art on the DeepFashion2 dataset.

Table~\ref{tab:retrieval-results-dresscode} reports Recall@K on DressCode for Proxynca++, Hyp-DINO, SigLIP2-FI2I, SigLIP2-Ensemble, and our Total-Ensemble.  
Among individual models, Hyp-DINO performs best, outperforming both Proxynca++ and SigLIP2-FI2I.  
Our Total-Ensemble, which combines SigLIP2-I2I, SigLIP2-T2I, Proxynca++, and Hyp-DINO, achieves 89.3\% R@1, 96.6\% R@5, and 98.0\% R@10.  
This is nearly 12 points higher at R@1 than the strongest single model (Hyp-DINO), highlighting the complementary strengths of SigLIP-based and metric learning approaches.

These results demonstrate that while SigLIP2-based retrieval is highly effective in a zero-shot setting.
Further gains are possible by combining it with specialized metric-learning models.
The ensemble strategy ensures robustness, mitigates outliers, and integrates diverse similarity signals, which is especially important for noisy, heterogeneous datasets such as our garment–lookbook collection.

\begin{table}[t]
    \centering

    \begin{tabularx}{\linewidth}{l *{3}{>{\centering\arraybackslash}X}}
        \toprule
        & R@1 & R@5 & R@10 \\
        \midrule
        SigLIP2-FI2I  (Ours)    & 67.7\% & 80.8\% & 84.9\% \\
        SigLIP2-T2I  (Ours)    & 63.6\% & 81.4\% & 86.2\% \\
        SigLIP2-I2I (Ours)     & 80.6\% & 91.6\% & 94.1\% \\
        Proxynca++       & 72.3\% & 87.4\% & 91.5\% \\
        Hyp-DINO         & 77.6\% & 90.2\% & 93.1\% \\
        Total-Ensemble  (Ours)  & \textbf{89.3\%} & \textbf{96.6\%} & \textbf{98.0\%} \\
        \bottomrule
    \end{tabularx}
        \caption{Retrieval performance (\%) for DressCode benchmark. The highest recall is marked in bold and the second best is marked with underline. Our model outperforms all previous models by over 10 percentage points.}
    \label{tab:retrieval-results-dresscode}
\end{table}

\subsection{Dataset}

Building on the strong performance of our Total-Ensemble retrieval model, we use it to construct garment–lookbook pairs from our image corpus.
With over 550,000 gallery images, retrieving matches for every query is computationally infeasible.
To address this, we limit retrieval to the top 2,000 most similar gallery images per query.
Because the four retrieval models often rank different lookbook images in their top matches, computing a simple mean similarity is not possible.
Instead the mean, we adopt the second-highest similarity score across the ensemble as a robust indicator.

\begin{figure}[t]
    \centering
    \begin{tikzpicture}
      \begin{axis}[
        xmode=log,
        xlabel={Index},
        ylabel={\% Matches},
        ymin=0, ymax=80,
        grid=both,
        width=\linewidth,
        height=5.5cm,
        tick label style={font=\small},
        label style={font=\small},
        legend style={font=\small},
      ]
        \addplot[
          mark=o,
          thick,
          color=blue
        ] coordinates {
          (100,71)
          (2000,58)
          (8000,50.5)
          (32000,42)
          (128000,35.5)
          (512000,25.5)
          (2048000,8)
        };
        \addlegendentry{\% Matches}
      \end{axis}
    \end{tikzpicture}
    \caption{
        Shows the garment retrieval accuracy of our dataset at indices 100, 2000, 8000, 32000, 128000, 512000, and 2048000, obtained with qualitative evaluation of 200 garment–lookbook image pair samples at each index, where the dataset is sorted by the similarity scores between the garment and lookbook image pairs.
    }
    \label{fig:qualitative-dataset-analysis}
\end{figure}
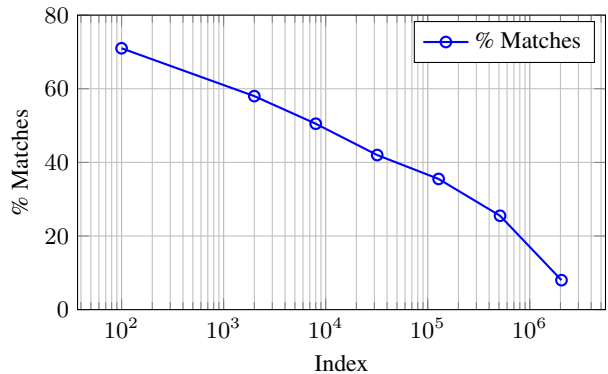

To assemble the dataset, we rank all garment images by their highest similarity match and select the top-K pairs for each quality tier.
We determine cutoffs based on Figure~\ref{fig:qualitative-dataset-analysis}, where we manually annotate 200 sampled pairs at indices 100, 2,000, 8,000, 32,000, 128,000, 512,000, and 2,048,000.
A pair is considered a true match only if the garments are visually indistinguishable, ensuring high fidelity for generative modeling.
Based on these observations, we divide the dataset into three quality tiers to support different experimental needs:
\textbf{High Quality:} 10,000 pairs suitable for precise evaluation or fine-tuning.
\textbf{Medium Quality:} 50,000 pairs, providing a larger set for training with moderate noise.
\textbf{Low Quality:} 300,000 pairs, enabling large-scale pretraining.

The dataset is designed for training diffusion models that generate lookbook images from garment inputs.  
High-quality pairs provide clean correspondences for fine-tuning, while medium and low-quality pairs, though noisier, increase diversity and scale to capture variations in poses, backgrounds, and editorial styles.  
By balancing fidelity and scale, the dataset supports robust training, enabling models to produce realistic, contextually grounded virtual photo-shoots from standardized garment images.

\section{Discussion and Conclusion}

We introduced the task of \textit{virtual photo-shoot}, aiming to generate editorial-style fashion imagery that goes beyond standard virtual try-on systems.
To support this, we constructed the first large-scale dataset of garment–lookbook pairs, bridging standardized product images with diverse, creative fashion visuals.
Unlike existing virtual try-on datasets, which primarily contain clean e-commerce product and shop-lookbook images, our dataset includes a variety of backgrounds, poses, editing styles, and creative compositions, enabling research into more context-rich fashion image generation.

Because such pairs are not naturally available, we developed an automated zero-shot retrieval pipeline combining SigLIP2-based similarity estimation (image-to-image and image-to-text), object-level reasoning, and vision–language alignment.
An ensemble of SigLIP2, Proxynca++, and Hyp-DINO further improved robustness and coverage across noisy, heterogeneous data, substantially outperforming individual models in recall@K across DeepFashion in-shop and consumer-to-shop, DeepFashion2, and DressCode benchmarks.

The dataset is organized hierarchically: high-quality pairs provide precise correspondences for fine-tuning generative models, while medium and low-quality pairs increase scale and diversity, crucial for training diffusion models to capture variations in poses, backgrounds, and styles.
This structure balances fidelity with coverage, supporting controlled, large-scale training and realistic, contextually grounded virtual photo-shoot generation.

Overall, our contributions highlight the potential of combining vision–language models, object detection, and metric learning for robust garment retrieval and dataset construction.
Looking ahead, future work could extend the dataset beyond luxury fashion to include smaller brands, refine retrieval through fine-grained attribute supervision, and leverage the dataset for generative modeling tasks such as controllable virtual photo-shoot synthesis.
By bridging the gap between e-commerce product imagery and creative fashion photography, we hope this work inspires new research at the intersection of computer vision, fashion, and generative modeling.

\bibliographystyle{IEEEbib}
\bibliography{strings,refs}

\end{document}